\pdfoutput=1

\documentclass[11pt]{article}

\usepackage{acl}
\usepackage{todonotes}

\usepackage{times}
\usepackage{latexsym}

\usepackage[T1]{fontenc}

\usepackage[utf8]{inputenc}
\usepackage{colortbl}

\usepackage{microtype}

\usepackage{inconsolata}

\usepackage{multirow}
\usepackage{booktabs}
\usepackage{graphicx}
\usepackage{tcolorbox}
\usepackage{xcolor}
\usepackage{bbm}
\usepackage{amsmath}
\usepackage{comment}
\usepackage{caption}
\usepackage{subcaption}
\usepackage[normalem]{ulem}
\useunder{\uline}{\ul}{}

\definecolor{ao(english)}{rgb}{0.0, 0.5, 0.0}

\newcommand{\sys}{{\itshape SyncTOD}}
\newcommand{\gptf}{{\itshape GPT-4}}
\newcommand{\gptc}{{\itshape ChatGPT}}

\newcommand{\llama}{{\itshape LLaMA2}}
\newcommand\prompt[1]{\texttt{#1}}

\title{Synergizing In-context Learning with Hints\\ for End-to-end Task-oriented Dialog Systems}

\author{Vishal Vivek Saley $^{1}$,  Rocktim Jyoti Das\thanks{*Work done when author was at IIT Delhi.} $^{3}$, Dinesh Raghu $^{2}$ and Mausam $^{1}$ \\
        $^{1}$ Indian Institute of Technology, Delhi\\
        $^{2}$ IBM Research, New Delhi, India\\
        $^{3}$ MBZUAI\\
        Vishal.Vivek.Saley@cse.iitd.ac.in, rocktimjyotidas@gmail.com \\ diraghu1@in.ibm.com, mausam@cse.iitd.ac.in}

\begin{document}
\maketitle
\begin{abstract}
End-to-end Task-Oriented Dialog (TOD) systems typically require extensive training datasets to perform well. In contrast, large language model (LLM) based TOD systems can excel even with limited data due to their ability to learn tasks through in-context exemplars. However, these models lack alignment with the style of responses in training data and often generate comprehensive responses, making it difficult for users to grasp the information quickly. In response, we propose \sys{} that synergizes LLMs with task-specific hints to improve alignment in low-data settings. \sys{} employs small auxiliary models to provide hints and select exemplars for in-context prompts. With \gptc{}, \sys{} achieves superior performance compared to LLM-based baselines and SoTA models in low-data settings, while retaining competitive performance in full-data settings.

\end{abstract}

\section{Introduction}

The rise of large-language models (LLMs) has progressed the field of NLP by leaps and bounds \citep{Anil2023PaLM2T,Touvron2023Llama2O}.
Pre-trained over massive data, LLMs work remarkably well with just in-context learning for many NLP tasks like natural language inference, summarization, and dialogs \citep{Kavumba2023PromptingFE,Hu2022InContextLF,vicuna_bench}. 

One specific domain within dialogs where LLMs show promise is in building Task-Oriented Dialogs (TOD) systems, where they generate agent responses based on the dialog history and task-specific knowledge. TOD systems, in general, can be divided into two types: modular \citep{Young2013POMDPBasedSS} and end-to-end \citep{Madotto2018Mem2SeqEI}. Modular systems require domain experts to define dialog states and annotate each train dialog with state annotations. Unlike modular, end-to-end systems do not require such expensive state annotations. In this work, we focus on end-to-end TOD systems.

\begin{table}[]
\centering
\resizebox{0.48\textwidth}{!}{
\begin{tabular}{lcccc}
\toprule
\fontsize{9pt}{9pt}\selectfont \fontsize{9pt}{9pt}\selectfont Name                   & \fontsize{9pt}{9pt}\selectfont Food    & ...          & \fontsize{9pt}{9pt}\selectfont Area   & \fontsize{9pt}{9pt}\selectfont Pricerange \\ \midrule
\fontsize{9pt}{9pt}\selectfont \fontsize{9pt}{9pt}\selectfont Graffiti               & \fontsize{9pt}{9pt}\selectfont British & \textbf{...} & \fontsize{9pt}{9pt}\selectfont west   & \fontsize{9pt}{9pt}\selectfont expensive  \\
\fontsize{9pt}{9pt}\selectfont \fontsize{9pt}{9pt}\selectfont The Oak Bistro         & \fontsize{9pt}{9pt}\selectfont British & \textbf{...} & \fontsize{9pt}{9pt}\selectfont centre & \fontsize{9pt}{9pt}\selectfont moderate   \\
\fontsize{9pt}{9pt}\selectfont \fontsize{9pt}{9pt}\selectfont The Copper Kettle      & \fontsize{9pt}{9pt}\selectfont British & \textbf{...} & \fontsize{9pt}{9pt}\selectfont centre & \fontsize{9pt}{9pt}\selectfont moderate   \\
\fontsize{9pt}{9pt}\selectfont \fontsize{9pt}{9pt}\selectfont Travellers Rest        & \fontsize{9pt}{9pt}\selectfont British & \textbf{...} & \fontsize{9pt}{9pt}\selectfont west   & \fontsize{9pt}{9pt}\selectfont expensive  \\
\fontsize{9pt}{9pt}\selectfont \fontsize{9pt}{9pt}\selectfont Saint Johns Chop House & \fontsize{9pt}{9pt}\selectfont British & \textbf{...} & \fontsize{9pt}{9pt}\selectfont west   & \fontsize{9pt}{9pt}\selectfont moderate   \\
\fontsize{9pt}{9pt}\selectfont \fontsize{9pt}{9pt}\selectfont Restaurant One Seven   & \fontsize{9pt}{9pt}\selectfont British & \textbf{...} & \fontsize{9pt}{9pt}\selectfont centre & \fontsize{9pt}{9pt}\selectfont moderate   \\
\end{tabular}
}

\resizebox{0.48\textwidth}{!}{
\begin{tabular}{p{0.1\textwidth} | p{0.38\textwidth}} 
\toprule
\fontsize{9pt}{9pt}\selectfont Context            & \fontsize{9pt}{9pt}\selectfont \textbf{User:} Are there any restaurants that serve proper British food in town? \\ \midrule
\fontsize{9pt}{9pt}\selectfont Gold               & \fontsize{9pt}{9pt}\selectfont Oh yes quite a few. Which part of town will you be dining in? \\
\fontsize{9pt}{9pt}\selectfont MAKER              & \fontsize{9pt}{9pt}\selectfont There are no British restaurants. Would you like to try a different cuisine? \\


\fontsize{9pt}{9pt}\selectfont Few-shot (\gptf{}) & \fontsize{9pt}{9pt}\selectfont Yes, there are quite a few restaurants that serve British food. One of them is The Oak Bistro located at 6 Lensfield Road, with the postcode cb21eg. Another option is The Copper Kettle at 4 Kings Parade City Centre, postcode cb21sj. Would you like more options? \\

\fontsize{9pt}{9pt}\selectfont \sys{} (\gptf{})    & \fontsize{9pt}{9pt}\selectfont Yes, there are quite a few. Do you have a preference for the price range or area? \\

\bottomrule
\end{tabular}}
\caption{\gptf{} lists many potential options and extraneous details instead of seeking user input and lacks alignment with the gold.}\label{tab:motdlg}
\end{table}

Existing end-to-end task-oriented dialogue (TOD) systems perform well when a reasonable number of training dialogues are available. However, in many practical scenarios, only a limited number of expert-curated training dialogues are available. Figure \ref{fig:low_data_all} showcases the performance of MAKER, a state-of-the-art (SoTA) end-to-end TOD model, on varying training data scales. When training data is limited, existing TOD approaches fail to learn the underlying task effectively, leading to a performance drop.

In contrast, large language models (LLMs) with in-context learning perform better than supervised models when the training dataset is limited. The inherent reasoning capabilities of LLMs help them learn the associated task with just a few examples. Unfortunately, LLM-based TOD systems do not align well with the language and style in the training dialogs, often generating overly comprehensive responses. This alignment is crucial, particularly in scenarios like in-car voice assistants, where responses must be concise and easily consumable without causing distraction.

As an illustrative example, see the responses generated by various models in Table \ref{tab:motdlg}. We see that \gptf{} is good at reasoning but lacks alignment in presenting information. When the gold seeks additional user input when posed with excessive options, \gptf{} tends to be overly comprehensive, listing many potential options and extraneous details. This verbosity, while informative, can hinder users from easily grasping the information. On the other hand, MAKER, a SoTA supervised approach, is well aligned with agent utterances in training but makes many mistakes in reasoning.

\vspace{0.5ex}
\noindent
\textbf{Contributions:} We propose \emph{Synergizing in-context learning with hints for TOD (\sys{})}, that aligns LLMs with the stylings of the available training data.
In particular, it trains auxiliary models to provide LLMs (accessed via an API) with hints (such as expected entity types in the response and response length) on how to phrase the response; selecting exemplars conditioned on these hints further improves the alignment of the responses. On three publicly available datasets, \sys{} consistently outperforms both vanilla prompting and SoTA supervised models in low-data settings while maintaining competitive performance compared to supervised models in full-data settings. 

\section{Related Work}
Conventional TOD systems follow a modular design \citep{Young2013POMDPBasedSS, RojasBarahona2016ANE, HosseiniAsl2020ASL, qin2023endtoend} and require annotations for DST, PL and NLG. This work, however, focuses on end-to-end TOD systems \citep{Eric2017KeyValueRN, Madotto2018Mem2SeqEI, raghu2018disentangling, Wu2019GlobaltolocalMP, qin2023endtoend} that alleviate the need for annotations by directly predicting the response given dialog history and knowledge base (KB).

Though LLMs have been explored for TOD tasks \citep{Hu2022InContextLF, Hudecek2023AreLA, Bang2023AMM, Li2023GuidingLL}, to the best of our knowledge, we are the first to explore them in an end-to-end setting. Directional Stimulus Prompting (DSP), an approach closer to ours, uses keywords and dialog acts as hints for summarization and response generation tasks, respectively \citep{Li2023GuidingLL}. However, unlike DSP, \sys{} uses multiple hints -- entity types, response length, and dialog closure -- relevant to the TOD task. Further, \sys{} also uses these hints to improve the in-context exemplars' quality using a retrieve-rerank approach.

A natural approach for combining training data with in-context learning is via retrieval-augmented generation (RAG) \citep{Lewis2020RetrievalAugmentedGF, Guu2020REALMRL}. Here, a retriever model infuses LLM input with exemplars from the training that are similar to the test sample \citep{Lewis2020RetrievalAugmentedGF, Meade2023UsingIL, Shi2023REPLUGRB, Ram2023InContextRL}. Although out-of-box retrievers work reasonably well \citep{Ram2023InContextRL}, many recent works strive to improve the retriever model further. \cite{Zhang2023RetrieveAT, Wang2023LearningTR} employ reward-based and contrastive learning to improve retrieval quality. Specifically, they use LLMs to obtain soft rewards to fine-tune the retriever model. Recently, \citet{patidar2024few} fused multiple retriever models learned from training data with LLMs for knowledge-based question-answering tasks.  What sets \sys{} apart from RAG is its use of hints not only for selecting the informative exemplars but also for steering LLM generation from within the prompt.

\section{\sys}\label{sec:approach}
\begin{figure*}[]
    \centering
    \includegraphics[width=0.99\textwidth]{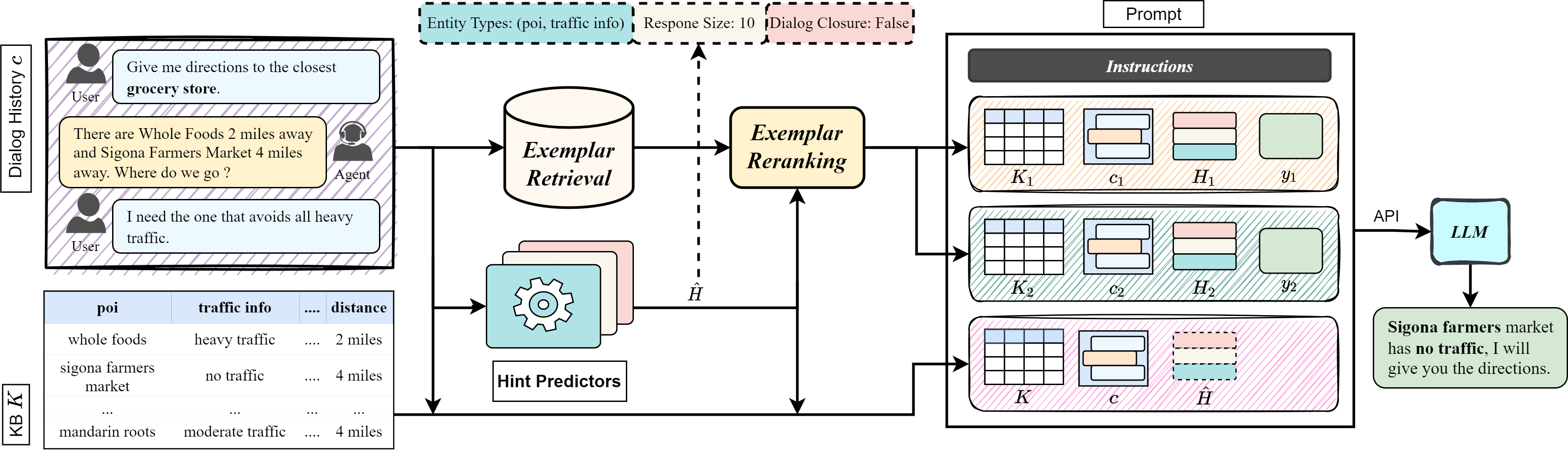}
    \caption{\sys{} predicts useful hints $\hat{H}$ about the expected response. The hints improve exemplar quality via re-ranking and steer the LLM (accessed via API) toward the expected response from within the prompt.}\label{fig:approach}
\end{figure*}

Let $c = [u_1, a_1, u_2, a_2,..., u_j]$ be a user-agent dialog history with $u$ and $a$ being user and agent utterances respectively. Let $y=a_j$ be the next system response. The task of a TOD system is to predict the next system response $\hat{y}$ given the dialog history $c$ and a knowledge base (KB) $K$ associated with the user's task. Let $\mathcal{D} = \{(h_i, K_i, y_i)\}_{i=1}^{n}$ denote the train dialogs.

%


In the in-context learning setup, an LLM is queried (via API) with an input prompt containing task instructions, a few exemplars, and $(c, K)$ to generate $\hat{y}$. A popular technique for leveraging train dialogs in the in-context learning setup is retrieval augmented generation (RAG) \citep{Zhang2023EnhancingPO, Guu2020REALMRL}. In RAG, the exemplars that are most similar to $c$ are retrieved from $\mathcal{D}$ and are used for generating $\hat{y}$.



Our proposed approach, \sys{}, synergizes in-context learning of LLMs with \emph{hints} to better align with agent utterances in the training data $\mathcal{D}$. Figure \ref{fig:approach} shows the overall architecture. \sys{} has two main components: hint predictors and exemplar selector. The hint predictors output a set of hints $\hat{H}$ given the dialog history $c$. These hints are domain-agnostic clues, such as the entity types that should be included in the response and the length of the response, that can guide the generation to follow the same style as the train dialogs. The second component, exemplar selector, first retrieves relevant exemplars from $\mathcal{D}$ based on $c$,  and then re-ranks the retrieved exemplars based on $\hat{H}$. Both these components are aimed at aligning the language and style of LLM responses to agent responses in the train dialogs $\mathcal{D}$. 
As the gold responses $y$ are available for the exemplars, we simply infer the corresponding hints from $y$ and add the hints to the exemplars.
The predictors are only used to infer hints for the given input dialog with history $c$.
Please refer to appendix \ref{section:prompt_spec} for the exact prompt. 



\subsection{Hint Predictors}
\sys{} uses three types of hints: entity types (in response), response length, and dialog closure.

\vspace{1ex}
\noindent \textbf{Entity Types (ET):} Entities are the information-rich elements in the agent's response. For example, the \textit{hotel} name "Lovell Lodge" is the crucial element in the agent response ``How does the Lovell Lodge sound?".
We posit that for a given dialog context and KB, the set of entity types in the agent response (e.g., \{\textit{hotel name}\}) captures the crux of the response. Hence using expected entity types in the response as hints would align the LLM generation to $\mathcal{D}$.

Specifically, for given $(c, K)$, \sys{} predicts a list of entity types $\hat{et}$ present in the expected system response. Then, \sys{} amends the prompt with the rule -- \textit{The response must only include entities of type: $\hat{et}$}. 
To predict $\hat{et}$, \sys{} learns an ET predictor model $P(et|c, K)$ on the dataset $\{(c_i,K_i,et_i)\}_{i=1}^n$, where gold $et_i$s are the types of entities in gold response.


\vspace{1ex}
\noindent \textbf{Dialog Closure (DC):} The style of the dialog closures varies depending on the task at hand, and each dataset has a different way of closing the dialog. But \gptc{} generates similar, verbose and open-ended responses to the user's closing salutations.
To alleviate this, \sys{} uses dialog closure prediction $dc$ for a given dialog $(c, K)$ as a hint to steer LLM towards a successful closure of the dialog. Specifically, \sys{} 
amends the input prompt with a rule: \textit{The response must close the dialog.}, when $dc$ is true.
For a training dialog $(c_i, K_i, y_i)$, we define $dc=\text{True}$ if and only if $y_i$ is the last utterance in the dialog.


\vspace{1ex}
\noindent \textbf{Response size (RS):} For a $(c_i, K_i, y_i) \in \mathcal{D}$, response size $rs$ equals the number of words in the response $y_i$. \sys{} learns an RS predictor $P(rs|c,K)$ on the dataset $\{(c_i,K_i,rs_i)\}_{i=1}^n$ and amends the input with rule: \textit{The response must be $rs$ words or shorter}.

For a test dialog $(c, K)$, \sys{} predicts the hints $\hat{H} = (\hat{et}, \hat{rs}, \hat{dc})$ using ET, RS, and DC hint predictors, respectively.

\subsection{Exemplar Selector}
\noindent \textbf{Retrieval:} \sys{} has a retrieve-rerank mechanism to select in-context exemplars \citep{Nogueira2019PassageRW}. Following \citet{Liu2021WhatMG}, \sys{} selects points from $\mathcal{D}$ semantically closer to the given test dialog $(c, K)$. Specifically, it encodes the dialog history $c$ using a pre-trained encoder and performs a maximum inner-product search over $\mathcal{D}$ to retrieve the top-$k$ points.
All our experiments use \textit{BAAI/bge-large-en-v1.5} encoder model \citep{bge_embedding}.

\vspace{1ex}
\noindent \textbf{Re-ranking:} Intuitively, an example with the same dialog state as the input is an ideal choice for an exemplar. However, end-to-end TOD datasets do not include dialog state annotations. Instead, we posit that dialog history and hints are reasonable proxies for the dialog state. \sys{} thus re-ranks the retrieved datapoints based on hints.

Let $(c_i, K_i, y_i)$ be a retrieved datapoint and $H_i$s be its associated hints. \sys{} computes similarity score between hints $\hat{H}$ and $H_i$ as follows
\begin{equation*}
f_h(\hat{H}, H_i) = 0.5 * \mathbbm{1}[\hat{dc} = dc_i] + 0.5 * \mathcal{J}(\hat{et}, et_i)
\end{equation*}
where $\mathbbm{1}$ is an indicator function and $\mathcal{J}$ is Jaccard similarity. From $k$ retrieved samples, \sys{} selects the top two with the highest hint similarity score as exemplars.

\section{Experimental Setup}\label{sec:experiments}
\noindent \textbf{Datasets} For our evaluation, we use the MultiWOZ2.1 \citep{Budzianowski2018MultiWOZA}, Stanford Multi-domain (SMD) \citep{Eric2017KeyValueRN}, and BiTOD (English) \citep{lin2021bitod} multi-domain datasets. Appendix \ref{sec:app_datasets} provides additional details about the datasets.

\vspace{0.5ex}
\noindent \textbf{Baselines:} We compare \sys{} against the recent baselines - GLMP \citep{Wu2019GlobaltolocalMP}, FG2Seq \citep{He2020Fg2seqEE}, CDNet \citep{Raghu2021ConstraintBK}, UnifiedSKG \citep{Xie2022UnifiedSKGUA}, and MAKER \citep{Wan2023MultiGrainedKR}. We also compare against RAG with \textit{BAAI/bge-large-en-v1.5} model for exemplar retriever. Further, we report the performance of \gptc{} \textit{(gpt-3.5-turbo)} and \gptf \textit{(gpt-4-0613)} in a standard few-shot setting with fixed exemplars\footnote{We set temperature $=$ 0 for LLMs generations.}. 
Training details
for hint predictors and retrieval of \sys{} are in Appendix~\ref{app:app_train_details}.


\vspace{0.5ex}
\noindent
\textbf{Evaluation Metric:} For evaluating model performance, we use the Entity F1 \citep{Wu2019GlobaltolocalMP} and BLEU \citep{Papineni2002BleuAM} metrics prevalent in the end-to-end TOD paradigm \citep[inter alia]{Wu2019GlobaltolocalMP, He2020Fg2seqEE, Raghu2021ConstraintBK, Xie2022UnifiedSKGUA, Wan2023MultiGrainedKR}.





\section{Results}\label{sec:results}

\begin{figure*}
    \centering
    \includegraphics[width=0.99\textwidth]{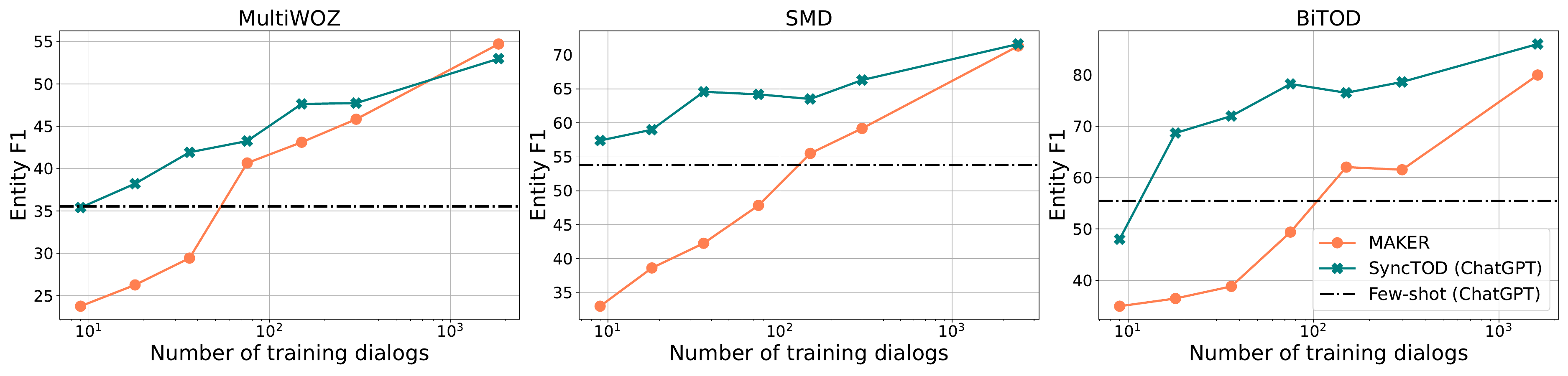}
    \caption{\sys{} performance across varying training data sizes.}\label{fig:low_data_all}
\end{figure*}


\noindent \textbf{Full Data Setting:} 
Table \ref{tab:main_results} summarizes the performance of various models under full-data setting. Across all datasets, \sys{} variants demonstrate competitive Entity F1 scores, with \sys{} (\gptf{}) outperforming all the supervised baseline models. Further, \gptc{} and \gptf{} enjoy consistent performance gains when coupled with \sys{}.

Interestingly, RAG LLMs display a stronger Entity F1 performance on SMD than other datasets. In SMD, users express preferences differently than the other two datasets. In MultiWOZ and BiTOD, users give detailed preferences for area, price, rating, etc., and can change these during the conversation. In SMD, preferences are simpler, like the nearest parking, city weather, or meeting times. Thus, MultiWOZ and BiTOD present a more challenging problem for LLMs than SMD.

Unlike Entity F1, \sys{} variants perform poorly in BLEU. Entity F1 measures whether the system response includes relevant entities from the KB and dialog history. Whereas BLEU computes n-gram precision between the system response and the gold response. Notably, a system response that includes all relevant entities, can still receive a low BLEU score due to differences in phrasing. We find that \sys{} responses are meaningful and include relevant entities, resulting in good Entity F1 scores. However, they use different phrasing and have less lexical overlap with gold responses, leading to lower BLEU scores. We verify the quality of \sys{} responses via human evaluations.


\begin{table}[]
\centering
\resizebox{0.5\textwidth}{!}{%
\begin{tabular}{l|cc|cc|cc}
\toprule
\multirow{2}{*}{Model} & \multicolumn{2}{c|}{MultiWOZ} & \multicolumn{2}{c|}{SMD} & \multicolumn{2}{c}{BiTOD}            \\
\cmidrule(r){2-7}
                      & BLEU       & Entity F1       & BLEU     & Entity F1    & \multicolumn{1}{c}{BLEU} & Entity F1 \\ \midrule
GLMP                  & 6.9            & 32.4           & 13.9           & 60.7           & 23.55          & 68.87          \\
FG2Seq                & 14.6           & 36.5           & 16.8           & 61.1           & 32.09          & 82.91          \\
CDNet                 & 11.9           & 38.7           & 17.8           & 62.9           & 25.49          & 77.13          \\
UnifiedSKG (T5-Large) & 13.69          & 46.04          & 17.27          & 65.85          & \textbf{36.73} & {\ul 88.62}    \\
MAKER (T5-Large)     & \textbf{18.77} & {\ul 54.72}    & \textbf{25.91} & 71.3           & 32.21          & 80.00          \\ \midrule
Zero-shot (\gptc{})   & 3.39           & 28.16          & 6.91           & 60.11          & 3.37           & 38.37          \\
Few-shot (\gptc{})    & 8.83           & 40.25          & 17.21          & 70.58          & 12.09          & 55.50          \\
Few-shot (\gptf{})       & 6.25           & 36.47          & 10.08          & 63.57          & 16.67          & 83.43          \\
RAG (\gptc{})         & 8.89           & 40.2           & 16.71          & 70.25          & 10.33          & 53.62          \\
RAG (\gptf{})           & 7.64           & 41.14          & 13.44          & 71.02          & 8.09           & 56.93          \\
\sys{} (\gptc{})     & 14.33          & 52.99          & 22.08          & {\ul 71.60}    & 19.81          & 86.04          \\
\sys{} (\gptf{})       & 13.01          & \textbf{54.99} & 19.08          & \textbf{72.99} & 19.34          & \textbf{89.04} \\ \bottomrule
\end{tabular}%
}
\caption{Performance of \sys{} and baselines on MultiWOZ, SMD and BiTOD datasets.}\label{tab:main_results}
\end{table}

\vspace{0.5ex}
\noindent \textbf{Human Evaluations:} We had two annotators evaluate responses from Gold, MAKER, 
\footnote{We used resources at \url{https://github.com/18907305772/MAKER} to obtain MAKER responses.},
and \sys{} (\gptf) models. They assessed the responses for a) \textit{appropriateness} to the dialog history and KB, b) \textit{fluency} and c) \textit{consistency} on a 1-5 Likert Scale \citep{Likert1932ATF}. The results in Table \ref{tab:human_evaluation} demonstrate that \sys{} surpasses MAKER in appropriateness and fluency across datasets, indicating higher response quality. Consistency evaluation showcases \sys{} is truthfulness to the dialog history and the KB. Appendix \ref{sec:app_human_eval} details our evaluation protocol.


\begin{table}[]
\centering
\small
\resizebox{0.49\textwidth}{!}{
\begin{tabular}{l|ccc|ccc}
\toprule
\multirow{2}{*}{Model}  & \multicolumn{3}{c|}{MultiWOZ} & \multicolumn{3}{c}{SMD} \\ \cmidrule(r){2-7}
                & Appro.        & Fluency  &  Consist.  & Appro.      & Fluency &  Consist. \\ \midrule
MAKER           & 4.38          & 4.79     &  4.29   & 4.51        & 4.97 & 4.57 \\
Gold            & 4.62          & \textbf{4.9}  &   4.51  & 4.79        & 4.95 & \textbf{4.8}  \\
\sys{} (\gptf{})  & \textbf{4.68} & 4.8  &  \textbf{4.74}  & \textbf{4.81}     & \textbf{4.98} &  4.71 \\ \bottomrule
\end{tabular}}
\caption{Human evaluation results.}\label{tab:human_evaluation}
\end{table}



\vspace{0.5ex}
\noindent \textbf{Low Data Setting:} Figure \ref{fig:low_data_all} shows the evaluation with varying training data sizes. \sys{} (\gptc) consistently enhances \gptc{} performance and outperforms MAKER with limited data. In MultiWOZ, \sys{} (\gptc) leads until MAKER catches up at around 1000 dialogs. In SMD, \sys{} (\gptc) achieves Entity F1 similar to MAKER with less than 20 examples, while MAKER needs 16x more data. In BiTOD, \sys{} (\gptc) significantly surpasses MAKER across training data scales.

\vspace{0.5ex}
\noindent \textbf{Alignment Study:}  \sys{} aligns LLM responses with the dataset style. We validate this by comparing the average response length (Avg Len) and average entity count (Avg Ent) of gold and \sys{} responses from the test set (Table \ref{tab:align}). \sys{} stats are closer to gold than RAG, indicating better alignment.

\begin{table}[]
\centering
\resizebox{0.48\textwidth}{!}{%
\begin{tabular}{l|cc|cc|cc}
\toprule
\multirow{2}{*}{\textbf{Model}} & \multicolumn{2}{c|}{MultiWOZ} & \multicolumn{2}{c|}{SMD} & \multicolumn{2}{c}{BiTOD} \\ \cmidrule{2-7}
 & Avg Len & Avg Ent & Avg Len & Avg Ent & Avg Len & Avg Ent \\ \midrule
Gold & 17.86 & 1.49 & 10.87 & 1.65 & 13.5 & 1.18 \\
RAG (\gptc) & 24.19 & 2.92 & 12.91 & 2.25 & 22.33 & 1.42 \\
\sys{} (\gptc) & \textbf{15.83} & \textbf{2.14} & \textbf{9.37} & \textbf{1.75} & \textbf{14.75} & \textbf{0.99} \\ \bottomrule
\end{tabular}%
}
\caption{\sys{} is better aligned with Gold than RAG.}
\label{tab:align}
\end{table}




\vspace{0.5ex}
\noindent \textbf{Ablations:} We perform ablations on \sys{} (\gptc{}), with results in Table \ref{tab:ablations}. Hints and exemplar retrieval are critical for \sys{}'s performance across datasets. Dropping exemplar re-ranking significantly impacts MultiWOZ and BiTOD but not SMD, likely due to SMD's simpler dialogs, which allow \sys{} to retrieve high-quality exemplars without re-ranking.

\begin{table}[]
\centering
\small
\resizebox{0.48\textwidth}{!}{
\begin{tabular}{rccc}
\toprule
\multicolumn{1}{c}{}                                  & MultiWOZ & SMD & BiTOD  \\
\midrule
\multicolumn{1}{l}{\sys{} (\gptc{})}                  & 52.99    & 71.60 & 86.03 \\
w\textbackslash{}o hint prediction                    & 40.2    & 70.25 & 53.62 \\
w\textbackslash{}o exemplar retrieval                 & 45.47    & 66.84 & 63.44 \\
w\textbackslash{}o exemplar reranking                 & 49.94    & 71.60 & 78.04 \\
\bottomrule
\end{tabular}}
\caption{Ablation Study: Entity F1 on MultiWOZ, SMD and BiTOD datasets} \label{tab:ablations}
\end{table}

\section{Conclusion}
We propose \sys{} that leverages LLMs for end-to-end TOD. Given a dialog history and KB, \sys{} obtains hints about the expected response using auxiliary models. It then uses predicted hints to retrieve quality exemplars and guide LLMs toward the desired response. With automatic/human evaluation, we showed that \sys{} outperforms the SoTA baseline models. Further, \sys{} showcases a strong performance in the low-data setting. 

\section*{Limitations}
It would be interesting to see how \sys{} benefits from advanced prompting techniques like chain-of-thought and self-consistency. Further, \sys{} is only tested on English datasets, though the model can easily be extended to different languages by its design. 
Additionally, \sys{} performance can further be improved by designing much more sophisticated hints.
Finally, \sys{} involves both training the hint prediction modules and prompting an LLM, resulting in the cost of using LLMs and training the model.

\section*{Ethics Statement}
In this work, we use OpenAI's \gptc{} and \gptf{} which are commercial LLMs whose training details are not publicly available. Thus, it is unclear whether these models have seen the datasets used in this work during their training. In our experiments, we benchmark Zero-shot (\gptc{}) on all the datasets and report the performance in table \ref{tab:main_results}. As zero-shot (\gptc{}) performs poorly, we believe that our datasets were not part of \gptc{}'s training set.

\section*{Acknowledgements}
This work is supported by IBM AI Horizons Network grant, grants by Google, Verisk, and Microsoft, an IBM SUR award and the Jai Gupta chair fellowship by IIT Delhi. Vishal is supported by a Google Fellowship. We thank the IIT Delhi HPC facility for its computational resources. We are grateful to Microsoft AFMR for supporting this work.

\bibliography{custom}

\appendix

\section{Dataset Details}\label{sec:app_datasets}
For MultiWOZ and SMD datasets, we use the versions of the dataset released by \citet{Wan2023MultiGrainedKR}.
We adapt BiTOD dataset \citep{lin2021bitod} to end-to-end setting by associating KB to the English dialogs available in the dataset.
\begin{table}[h!]
\resizebox{0.5\textwidth}{!}{
\begin{tabular}{lllll}
\toprule
Dataset  & Domain                        & \#train & \#val & \#test \\
\midrule
MultiWOZ & Restaurant, Hotel, Attraction & 1839    & 117   & 141    \\
SMD      & Navigate, Schedule, Weather   & 2425    & 302   & 304    \\
BiTOD    & Restaurant, Hotel, Attraction & 1614    & 169   & 152    \\
\bottomrule
\end{tabular}}
\caption{Evaluation Dataset Details}
\end{table}

\section{Rules Ablation Study}\label{app:fine_grained_ablations}
We conducted an ablation study using the MultiWOZ dataset by removing individual hints from SyncTOD. The results are in the table \ref{tab:rule_abl}. Each hint is crucial for \sys{} performance, especially the entity types hint, whose removal significantly lowers performance.

\begin{table}[]
\centering
\resizebox{0.25\textwidth}{!}{%
\begin{tabular}{rc}
\toprule
\textbf{Configuration} & \textbf{Entity F1} \\ \midrule
SyncTOD & 52.99 \\
w\textbackslash{}o Entity Types & 41.85 \\
w\textbackslash{}o Dialog Closure & 51.38 \\
w\textbackslash{}o Response Length & 49.23 \\ \bottomrule
\end{tabular}%
}
\caption{Rules ablation results on MultiWOZ dataset.}\label{tab:rule_abl}
\end{table}

\section{Additional Baselines}\label{sec:add_baselines}
We compared our model against the following end-to-end TOD baselines - We compare \sys{} against the following baselines - DSR \citep{Wen2018SequencetoSequenceLF}, KB-Retriever \citep{Qin2019EntityConsistentET}, GLMP \citep{Wu2019GlobaltolocalMP}, DF-Net \citep{Qin2020DynamicFN}, GPT-2+KE \citep{Madotto2020LearningKB}, EER \citep{He2020TaskOrientedDG}, FG2Seq \citep{He2020Fg2seqEE}, CDNet \citep{Raghu2021ConstraintBK}, GraphMemDialog \citep{Wu2022GraphMemDialogOE}, ECO \citep{Huang2022AutoregressiveEG}, DialoKG \citep{Rony2022DialoKGKA}, UnifiedSKG \citep{Xie2022UnifiedSKGUA}, Q-TOD \citep{Tian2022QTODAQ} and MAKER \citep{Wan2023MultiGrainedKR}. Results are shown in table \ref{tab:main_results2}.

\begin{table}[h]
\centering
\resizebox{0.48\textwidth}{!}{
\begin{tabular}{l|cc|cc}
\toprule
\multirow{2}{*}{Model}     & \multicolumn{2}{c|}{MultiWOZ} & \multicolumn{2}{c}{SMD} \\
\cmidrule(r){2-5}
                           & BLEU        & Entity F1      & BLEU     & Entity F1    \\ \midrule
DSR                        & 9.1         & 30             & 12.7     & 51.9         \\
KB-Retriever               & -           & -              & 13.9     & 53.7         \\
GLMP                       & 6.9         & 32.4           & 13.9     & 60.7         \\
DF-Net                     & 9.4         & 35.1           & 14.4     & 62.7         \\
GPT-2+KE                   & 15.05       & 39.58          & 17.35    & 59.78        \\
EER                        & 13.6        & 35.6           & 17.2     & 59           \\
FG2Seq                     & 14.6        & 36.5           & 16.8     & 61.1         \\
CDNet                      & 11.9        & 38.7           & 17.8     & 62.9         \\
GraphMemDialog             & 14.9        & 40.2           & 18.8     & 64.5         \\
ECO                        & 12.61       & 40.87          & -        & -            \\
DialoKG                    & 12.6        & 43.5           & 20       & 65.9         \\
UnifiedSKG (T5-Large)      & 13.69       & 46.04          & 17.27    & 65.85        \\
Q-TOD (T5-Large)           & 17.62       & 50.61          & 21.33    & 71.11        \\
MAKER (T5-large)           & \textbf{18.77}       & 54.72          & \textbf{25.91}    & 71.3 \\ \midrule
Zero-shot (\gptc{})        & 3.39        & 	28.16   &   6.91    &   60.11 \\
Few-shot (\gptc{})         & 8.83        & 40.25          & 17.21    & 70.58        \\
Few-shot (\gptf{})         & 6.25        & 36.47          & 10.08    & 63.57        \\
RAG (\gptc{})           & 8.98       & 40.2          & 16.71    & 70.25        \\
RAG (\gptf{})           & 7.64       & 41.14         & 13.44    & 71.02        \\
Few-shot (\llama{} 70B)     & 5.26        &	39.68         &	  3.29    &   	46.20 \\
Few-shot (\llama{} Chat 70B) & 3.34	     &  30.33	& 3.15	& 53.27 \\
\sys{} (\llama{} 70B)        & 14.44	     & 50.51   &	15.37  &  63.33 \\
\sys{} (\llama{} Chat 70B)   & 8.35        &   48.01   &   7.92    &   63.31   \\
\sys{} (\gptc{})           & 14.33       & 52.99          & 22.08    & 71.60        \\
\sys{} (\gptf{})           & 13.01       & \textbf{54.99}          & 19.08    & \textbf{72.99}        \\
\midrule
\end{tabular}}
\caption{Performance of \sys{} and baselines on MultiWOZ and SMD datasets.}\label{tab:main_results2}
\end{table}

\section{Training \sys{} with Full Training Set}\label{app:app_train_details}
We use Nvidia V100 GPUs to train all our models. \\

\noindent \textbf{ET Predictors:} We model all the ET predictors as \textit{flan-t5-large} \citep{Chung2022ScalingIL} sequence predictors and train them for $8$ epochs with a learning rate (LR) of $1e-4$ and batch size (BS) of $32$. We use a linear decay LR scheduler with a warm-up ratio of $0.1$. We use AdamW optimizer \citep{Loshchilov2017DecoupledWD}. Training time was around 10 hours.

\vspace{1ex}
\noindent \textbf{DC Predictors:} We model all the DC predictors as \textit{deberta-V3-base} \citep{he2021debertav3} binary classifiers and train them for $5$ epochs with an LR of $3e-5$, BS of $16$, and linear decay LR scheduler with a warm-up ratio of $0.1$. We use AdamW optimizer. Training time was around 1 hour.

\vspace{1ex}
\noindent \textbf{RS Predictors:} During our experiments, we found that the training RS predictor is unstable. Thus, we use a constant RS predictor with a value equal to the mean response size in training data.

\vspace{1ex}
\noindent \textbf{Exemplar Retrieval:} For the MultiWOZ dataset, we use the last user utterance in the dialog context to dense retrieve $k=30$ samples from the training data. We then re-rank them based on the hints and pick the top two.

For the SMD dataset, we found that retrieval using the entire dialog context works the best. We attribute it to shorter dialog context and utterances in the SMD dataset. Further, we use $k=2$ as exemplars are already of high quality.

\section{Hint Predictors Performance}

\begin{table}[h]
\centering
\resizebox{0.48\textwidth}{!}{
\begin{tabular}{lccc}
\hline
Accuracy & MultiWOZ & SMD & BiTOD \\ \hline
Closure Prediction & 0.9564 & 0.9109 & 0.9570 \\ 
Entity Type Prediction & 0.6805 & 0.7436 & 0.8778 \\ \hline
\end{tabular}}
\caption{Accuracy of hint Predictor models.}
\label{tab:hint-pred-acc}
\end{table}

Table \ref{tab:hint-pred-acc} reports the performance of \sys{} hint predictors. We report accuracy for the DC predictor and micro F1 for the ET predictor. 
We observe that the DC predictor achieves high performance across datasets. However, ET predictors still show room for improvement, which indicates \sys{} performance can be pushed further.

\section{Human Evaluation Details}\label{sec:app_human_eval}
\begin{figure*}
    \centering
    \includegraphics[width=0.95\textwidth]{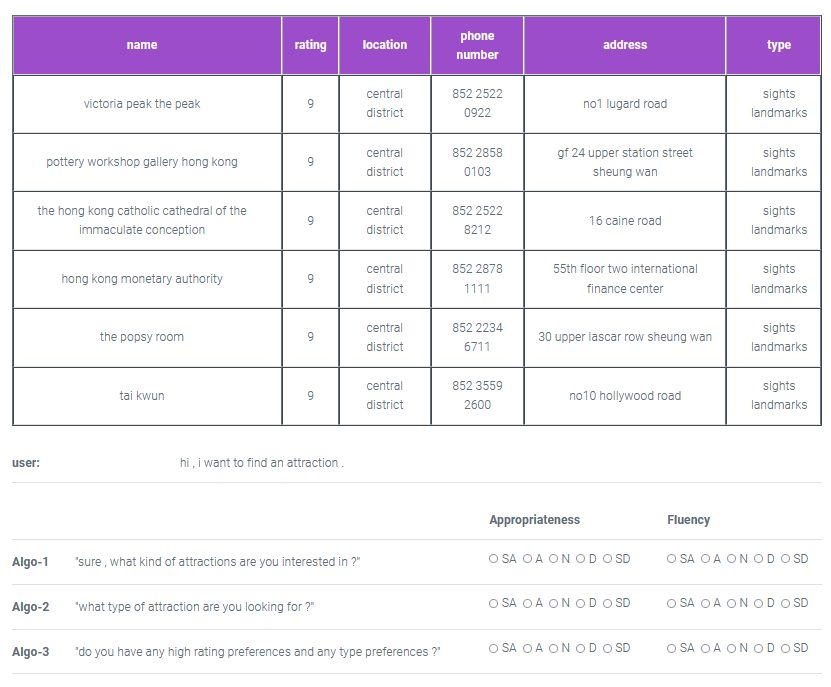}
    \caption{Human Evaluation Portal}\label{fig:portal}
\end{figure*}

A snapshot of our human evaluation portal is given in figure \ref{fig:portal}. Detailed evaluation guidelines are given at the end of this section. 

We human-evaluate responses from three TOD systems - Gold, MAKER, and \sys{} (\gptf). From MultiWOZ and SMD datasets, we sample 80 context-response pairs to evaluate appropriateness and fluency. Two annotators, undergraduate and graduate student volunteers, then independently score TOD system responses for these samples on a Likert scale \cite{Likert1932ATF}. Here, the inter-annotator agreement was Kendall's Tau $\tau = 0.47$ at $(p<0.0001)$.

To evaluate consistency, we randomly sample 60 context-response pairs from the two datasets. Two student volunteers rated responses from the Gold, MAKER, and SyncTOD systems on a 1-5 point Likert scale. One volunteer is a PhD scholar, while the other is a graduate student with a background in machine learning and NLP. Here, the inter-annotator agreement was Kendall's Tau $\tau = 0.45$ at $(p<0.0001)$.

The detailed evaluation guidelines are given below.

\par\noindent\rule{0.48\textwidth}{0.4pt}
\noindent
\textbf{Task Overview} \\
 
There are several dialog context response pairs in the html file. Each context response pair dictates a scenario where user is enquiring the agent about hotels, restaurant and attractions to visit. 
\begin{itemize}
    \item User can optionally request for additional attributes like phone number and address and can make a booking. 
    \item Agent is expected to suggest hotel, restaurant and attraction with the highest rating among available options. 
    \item     In each scenario, agent re-confirms details like user’s name, selected hotel/restaurant/attraction, number of people, rooms and dates before making the final booking. 
\end{itemize}

Along with the context response pair, there are outputs of different dialog systems (randomly shuffled). You are requested to annotate each system generated output along two dimensions: appropriateness and fluency using the following scale: 

\begin{enumerate}
    \item SA: Strongly Agree 
    \item A : Agree 
    \item N : Neutral 
    \item D : Disagree 
    \item SD: Strongly Disagree 
\end{enumerate}

\noindent
\textbf{How to judge appropriateness?}\\
\begin{enumerate}
    \item Strongly Agree - when the generated output conveys the intended information –correct entity (hotel/restaurant/attraction) and its attributes (address, phone, rating, etc). Also, when generated output requests correct input from the user. 
    \item Agree – when generated output contains partial information (e.g., when user request address and phone number but output contains only address). 
    \item Neutral – when generated output is hard to decide whether its right or wrong. 
    \item Disagree - when the generated response is somewhat unacceptable (e.g., re-querying already known information like cuisine for restaurants and name of the user for booking). 
    \item Strongly Disagree – when the generated output contains incorrect information (entities or attributes) for given conversation context. 
\end{enumerate}

\noindent
\textbf{How to judge fluency?}\\

Evaluate the linguistic quality of the response, including grammar, coherence, and readability.
The fluency of the response is independent of the dialog context or ground truth. 
A system output can be marked strongly disagree for appropriateness and still be marked strongly agree for fluency.
You can make your own rules about what each rating in the scale means for fluency, but please be consistent with the rules you come up with.
\par\noindent\rule{0.5\textwidth}{0.4pt}

\noindent
\textbf{How to judge Consistency?}\\

Consistency of system response is the degree to which the system's response accurately reflects and logically aligns with the dialogue history and the knowledge base. Please rate each system response on the following scale.

\begin{itemize}
    \item Strongly Disagree (SD): The response is completely inconsistent with the dialogue history and the knowledge base. It provides incorrect information, contradicts previous dialog, and does not align with known facts.
    \item Disagree (D): The response has significant inconsistencies with the dialogue history and knowledge base. It may provide some correct information but contains major errors or contradictions.
    \item Neutral (N): The response is generally consistent with the dialogue history and knowledge base but may include minor errors or inconsistencies. The response mostly aligns with the previous context but might have inaccuracies or ambiguities.
    \item Agree (A): The response is consistent with the dialogue history and knowledge base. It correctly addresses the context and facts, with only minor issues that do not significantly impact the overall coherence.
    \item Strongly Agree (SA): The response is fully consistent with the dialogue history and knowledge base. It accurately reflects the context, aligns perfectly with known facts, and shows no contradictions or irrelevant information.
\end{itemize}

\par\noindent\rule{0.5\textwidth}{0.4pt}

\subsection{BiTOD Human Evaluation}
We evaluate the MAKER, Gold, and \sys{} (\gptf) systems on the BiTOD dataset for Appropriateness and Fluency. The results, shown in Table \ref{tab:bitod_human_eval}, indicate that \sys{} outperforms MAKER in terms of Appropriateness, while all models demonstrate strong performance in Fluency.

\begin{table}[]
\centering
\resizebox{0.49\textwidth}{!}{%
\begin{tabular}{lcc}
\toprule
Model & Appropriateness & Fluency \\ \midrule
MAKER & 4.71 & \textbf{4.97} \\
Gold & 4.89 & 4.85 \\
\sys{} (\gptf) & \textbf{4.94} & 4.94 \\ \bottomrule
\end{tabular}%
}
\caption{Human evaluation results on BiTOD dataset.}
\label{tab:bitod_human_eval}
\end{table}

\section{Case Studies}
We present examples from MultiWOZ (table \ref{tab:woz_exp1}), SMD (tables \ref{tab:smd_exp1} and \ref{tab:smd_exp2}) and BiTOD (table \ref{tab:bitod_exp}) comparing responses from \sys{}, MAKER and the gold systems.

\begin{table*}[]
\centering
\resizebox{0.98\linewidth}{!}{
\begin{tabular}{llllllllllll}
\toprule
Name                        & Address                & Area   & Phone       & Postcode & Pricerange & Stars  & Type       & Choice & Ref      & Parking & Internet \\ \midrule
alpha-milton guest house    & 63 milton road         & north  & 01223311625 & cb41xa   & moderate   & 3 star & guesthouse & -      & epwuawm5 & no      & no       \\
alexander bed and breakfast & 56 saint barnabas road & centre & 01223525725 & cb12de   & cheap      & 4 star & guesthouse & -      & epwuawm5 & yes     & yes      \\
acorn guest house           & 154 chesterton road    & north  & 01223353888 & cb41da   & moderate   & 4 star & guesthouse & -      & epwuawm5 & yes     & yes      \\
allenbell                   & 517a coldham lane      & east   & 01223210353 & cb13js   & cheap      & 4 star & guesthouse & -      & epwuawm5 & yes     & yes      \\
bridge guest house          & 151 hills road         & south  & 01223247942 & cb28rj   & moderate   & 3 star & guesthouse & -      & epwuawm5 & yes     & yes      \\
a and b guest house         & 124 tenison road       & east   & 01223315702 & cb12dp   & moderate   & 4 star & guesthouse & -      & epwuawm5 & no      & yes     
\end{tabular}
}

\resizebox{0.98\linewidth}{!}{

\begin{tabular}{p{0.1\linewidth} | p{0.88\linewidth}}
\toprule
\fontsize{9pt}{9pt}\selectfont Context           & \fontsize{9pt}{9pt}\selectfont \textbf{User:} I' m looking for a place. I think it's called like bridge house or bridge guesthouse or something. \\
\fontsize{9pt}{9pt}\selectfont Gold              & \fontsize{9pt}{9pt}\selectfont You are looking for the Bridge Guest House. \\
\fontsize{9pt}{9pt}\selectfont MAKER             & \fontsize{9pt}{9pt}\selectfont I'm sorry, I don't have any information on that. Would you like to try a different name? \\
\fontsize{9pt}{9pt}\selectfont \sys{} (\gptc{})  & \fontsize{9pt}{9pt}\selectfont Bridge guest house is a moderate priced guesthouse in the south with a 3 star rating. Can I help you book a room there? \\
\fontsize{9pt}{9pt}\selectfont \sys{} (\gptf{})    & \fontsize{9pt}{9pt}\selectfont Yes, the Bridge Guest House is a 3 star guesthouse in the south with moderate pricing. \\ \bottomrule
\end{tabular}
}
\caption{\sys{} models understand user's requirement for \textit{bridge guest house} and present required information from the KB. MAKER, however, produces incorrect results.  } \label{tab:woz_exp1}
\end{table*}

\begin{table*}[]
\centering
\resizebox{0.98\linewidth}{!}{

\begin{tabular}{llllll}
\toprule
\textbf{Event}          & \textbf{Room}       & \textbf{Agenda} & \textbf{Time} & \textbf{Date} & \textbf{Party} \\ \midrule
conference              & conference room 102 & go over budget  & 7pm           & monday        & boss           \\
lab appointment         & -                   & -               & 9am           & tuesday       & tom            \\
dinner                  & -                   & -               & 7pm           & tuesday       & martha         \\
optometrist appointment & -                   & -               & 5pm           & the 5th       & brother       
\end{tabular}

}

\resizebox{0.98\linewidth}{!}{

\begin{tabular}{p{0.1\linewidth} | p{0.88\linewidth}}
\toprule

\fontsize{9pt}{9pt}\selectfont \multirow{3}{*}{Context} & \fontsize{9pt}{9pt}\selectfont \textbf{User:} I need to know my conference agenda, date and who is attending. \\
                         & \fontsize{9pt}{9pt}\selectfont \textbf{System:} Your conference is set for Monday with your boss to go over budget. \\
                         & \fontsize{9pt}{9pt}\selectfont \textbf{User:} Great, and who all will be there? \\
\fontsize{9pt}{9pt}\selectfont Gold                     & \fontsize{9pt}{9pt}\selectfont The conference is with your boss. \\
\fontsize{9pt}{9pt}\selectfont MAKER                    & \fontsize{9pt}{9pt}\selectfont You have three conference agendas, one for Monday with your boss to go over budget, one for Friday with your boss to go over the budget, and one for Monday with your boss to go over the budget. \\
\fontsize{9pt}{9pt}\selectfont \sys{} (\gptc{})        & \fontsize{9pt}{9pt}\selectfont Your boss will be attending the conference. \\
\fontsize{9pt}{9pt}\selectfont \sys{} (\gptf{})          & \fontsize{9pt}{9pt}\selectfont Your boss will attend the conference. \\                              

\bottomrule
\end{tabular}

}
\caption{\sys{} responses are grounded into KB while MAKER provides repeated and incorrect information.} \label{tab:smd_exp1}
\end{table*}


\begin{table*}[]
\centering
\resizebox{0.98\linewidth}{!}{

\begin{tabular}{lllll}
\toprule
Poi                        & Distance & Address             & Poi Type            & Traffic Info     \\ \midrule
Coupa                      & 6 miles  & 394 van ness ave    & coffee or tea place & moderate traffic \\
Mandarin Roots             & 2 miles  & 271 springer street & chinese restaurant  & heavy traffic    \\
Four Seasons               & 3 miles  & 465 arcadia pl      & rest stop           & no traffic       \\
Ravenswood Shopping Center & 1 miles  & 434 arastradero rd  & shopping center     & moderate traffic
\end{tabular}

}

\resizebox{0.98\linewidth}{!}{

\begin{tabular}{p{0.1\linewidth} | p{0.88\linewidth}}
\toprule

\fontsize{9pt}{9pt}\selectfont \multirow{3}{*}{Context} & \fontsize{9pt}{9pt}\selectfont \textbf{User:} Give me directions to the local starbuck  \\
                         & \fontsize{9pt}{9pt}\selectfont \textbf{System:} I do not have info on the local starbucks but i do see another coffee place. Would you like to hear info on it? \\
                         & \fontsize{9pt}{9pt}\selectfont \textbf{User:} Yes, please! Which one is its address? There is any quick route available?  \\
\fontsize{9pt}{9pt}\selectfont Gold                     & \fontsize{9pt}{9pt}\selectfont There is Coupa located at 394 van ness ave, there is moderate traffic but you're just 6 miles away \\
\fontsize{9pt}{9pt}\selectfont MAKER                    & \fontsize{9pt}{9pt}\selectfont I'm sorry, I do not have that info on you, I sent it on your screen, drive carefully! \\
\fontsize{9pt}{9pt}\selectfont \sys{} (\gptc{})        & \fontsize{9pt}{9pt}\selectfont Coupa is located at 394 van ness ave. There is moderate traffic on the way. \\
\fontsize{9pt}{9pt}\selectfont \sys{} (\gptf{})          & \fontsize{9pt}{9pt}\selectfont Coupa at 394 van ness ave has moderate traffic. \\
\bottomrule
\end{tabular}

}
\caption{\sys{} models accurately answer user's query.} \label{tab:smd_exp2}
\end{table*}

\begin{table*}[]
\centering
\resizebox{0.98\linewidth}{!}{

\begin{tabular}{lllllllll}
\toprule
{} &                         name &                                            address &  cuisine & dietary\_restrictions &               location &   phone\_number & price\_level & rating \\
\midrule
0 &             chocoduck\_bistro &                         3\_kin\_san\_lane\_cheung\_chau &  italian &        vegan\_options &        new\_territories &  852\_9735\_0049 &    moderate &     10 \\
1 &    la\_table\_french\_brasserie &  72\_mody\_road\_1f\_new\_world\_millennium\_hong\_kong... &   french &        vegan\_options &          tsim\_sha\_tsui &  852\_2313\_4222 &    moderate &     10 \\
2 &                 fleur\_de\_sel &  shop\_2j\_po\_foo\_building\_foo\_ming\_street\_causew... &   french &        vegan\_options &  wan\_chai\_causeway\_bay &  852\_2805\_6678 &    moderate &     10 \\
3 &                     serendib &  shop\_2\_gf\_nam\_wo\_hong\_building\_148\_wing\_lok\_st... &    asian &        vegan\_options &             sheung\_wan &  852\_3705\_2429 &    moderate &     10 \\
4 &                bombay\_dreams &                                  77\_wyndham\_street &   indian &        vegan\_options &                central &  852\_2971\_0001 &    moderate &      9 \\
5 &   divino\_wine\_bar\_restaurant &                          73\_wyndham\_street\_central &  italian &        vegan\_options &                central &  852\_2167\_8883 &    moderate &      9 \\
6 &  enchanted\_garden\_restaurant &  hong\_kong\_disneyland\_resort\_disneyland\_hotel\_l... &  seafood &        vegan\_options &        new\_territories &  852\_3510\_6000 &    moderate &      9 \\
7 &                      jw\_cafe &  5f\_jw\_marriott\_hotel\_hong\_kong\_pacific\_place\_n... &     cafe &        vegan\_options &       hong\_kong\_island &  852\_2810\_8366 &    moderate &      9 \\
\bottomrule
\end{tabular}

}

\resizebox{0.96\linewidth}{!}{

\begin{tabular}{p{0.1\linewidth} | p{0.86\linewidth}}
\toprule

\fontsize{9pt}{9pt}\selectfont \multirow{3}{*}{Context} & \fontsize{9pt}{9pt}\selectfont \textbf{User:} Hello. Can you help me find a place to eat ?  \\
                         & \fontsize{9pt}{9pt}\selectfont \textbf{System:} Would you like to choose the rating value for the restaurant? \\
                         & \fontsize{9pt}{9pt}\selectfont \textbf{User:} Not really . I just want to make sure they have vegan\_options .  \\
                         & \fontsize{9pt}{9pt}\selectfont \textbf{System:} Did you have a price range or cuisine type in mind ?  \\
                         & \fontsize{9pt}{9pt}\selectfont \textbf{User:} Any cuisine will work , but it needs to be moderately priced . \\
                         & \fontsize{9pt}{9pt}\selectfont \textbf{System:} I've found \#210 options with those criteria . May i recommend an Italian place with a rating of 10 called chocoduck\_bistro ? \\
                         & \fontsize{9pt}{9pt}\selectfont \textbf{User:} That sounds good . can i make a reservation ? \\
                         & \fontsize{9pt}{9pt}\selectfont \textbf{System:} Definitely ! how many will be in your party and what time would you like to book ? \\
                         & \fontsize{9pt}{9pt}\selectfont \textbf{User:} There will be 14 of us , and we can be there by 410 in the afternoon on Sunday . \\
                         & \fontsize{9pt}{9pt}\selectfont \textbf{System:} Wonderful , and what name should i put that under ? \\
                         & \fontsize{9pt}{9pt}\selectfont \textbf{User:} Put it under Danielle , please . \\
\fontsize{9pt}{9pt}\selectfont Gold                     & \fontsize{9pt}{9pt}\selectfont Great ! Just to confirm your reservation for Sunday at 410 in the afternoon at chocoduck\_bistro for a party of 14 , correct ? \\
\fontsize{9pt}{9pt}\selectfont MAKER                    & \fontsize{9pt}{9pt}\selectfont Okay , and what is your booking time ? \\
\fontsize{9pt}{9pt}\selectfont \sys{} (\gptc{})        & \fontsize{9pt}{9pt}\selectfont Got it, I can make a reservation for 14 people at chocoduck\_bistro on Sunday at 410 pm under the name Danielle. \\
\fontsize{9pt}{9pt}\selectfont \sys{} (\gptf{})          & \fontsize{9pt}{9pt}\selectfont So, a reservation for 14 at chocoduck\_bistro, 410 pm on Sunday. \\
\bottomrule
\end{tabular}

}
\caption{\sys{} models assists user in making the reservation.} \label{tab:bitod_exp}
\end{table*}

\section{Prompt Specification}\label{section:prompt_spec}
\subsection{Design}
\sys{} prompts are comprised of \prompt{instructions} followed by tuples \prompt{(database, rule, dialog, follow-up response)} for exemplars and test sample.

\vspace{1ex}
\noindent \prompt{instructions} - Task definitions and ontology details for the dataset.

\vspace{1ex}
\noindent \prompt{database} - KB $K$ associated with a sample (exemplar or test). We use JSON index format which we found to perform well during our seed experiments.

\vspace{1ex}
\noindent \prompt{rules} - We include hints $H$ as a set of rules in the prompt and ask the LLM to follow the rules for writing the response. Rules guide the LLM toward the desired answer. We provide further details on rule creation at the end of this section.

\vspace{1ex}
\noindent \prompt{dialog history} - User and system utterances in the dialog context $c$.

\vspace{1ex}
\noindent \prompt{follow-up response} - For exemplars, we succinctly re-iterate the task definition and the entity types expected in the response, followed by gold entities and the response. For the test sample, we only provide task definition and entity types expected in the response and prompt the LLM to generate entities and the final response in order.

\subsection{Creating rules from hints}
We transform hints $H=(et, dc, rs)$ to rules in the prompt as follows. For response size, We add a rule \prompt{The response must be $rs$ words or shorter}.
For dialog closure $dc = \text{True} (\text{False})$, we add a rule \prompt{The response must (not) close the dialog.}.
For entity types $et=[t_1, t_2, t_3]$, we add a rule \prompt{The response must only include entities of type - $t_1, t_2, t_3$.}. We also introduce a rule \prompt{The response must not include any entities of type - $t'_1, t'_2,..$} where $t'$ are entity types not present in $et$. We find that explicitly presenting negative entity types demotivates LLM from including extraneous entities in the response.

\clearpage
\clearpage
\subsection{Sample Prompts}\label{section:prompts}
\subsection*{MultiWOZ}\label{prompt:multiwoz}
\begin{tcolorbox}[
    colback=white,colframe=black,boxrule=0.25mm,
    width=\textwidth,
]
\fontsize{5pt}{5pt}\selectfont
Henceforth, assume that you are a customer support expert. I will give you an incomplete dialog between a user and a customer service representative. As an expert, you must suggest the most appropriate follow-up response to the dialog. Ensure you also include correct information (entities) from the given database. Entities can be of the following types - \newline
1. name - name of a place (restaurant, hotel or attraction)\newline
2. address - address of the place\newline
3. phone - phone number of the place\newline
4. food - the type of food a restaurant serves\newline
5. area - a region of the city, e.g. centre, north, south, east, west\newline
6. postcode - postcode of the place\newline
7. price range - price range of the place, e.g. cheap, moderate, expensive\newline
8. type - the type of a place, e.g. restaurant, hotel, gusesthouse, attraction\newline
9. reference number - reference code for booking, e.g. 542j9wog\newline
10. stars - star rating of the hotel, e.g. 3 stars\newline
11. choice - number of available choices that match user's requirements, e.g. many, few, several, 10\newline
\newline
As an expert, you are very strict about following rules. Make sure that the follow-up response you write follows all the given rules. Here are the examples -\newline
\newline
[example 1]\newline
[database 1]\newline
\{\newline
\hspace*{5pt}  "magdalene college":\{\newline
\hspace*{10pt}  "address":"magdalene street",\newline
\hspace*{10pt}  "phone":"01223332138",\newline
\hspace*{10pt}  "area":"west",\newline
\hspace*{10pt}  "postcode":"cb30ag",\newline
\hspace*{10pt}  "price range":"free",\newline
\hspace*{10pt}  "type":"college",\newline
\hspace*{10pt}  "choice":"79"\newline
\hspace*{5pt}  \},....\newline
\}\newline
\newline
[rules 1]\newline
The response must be 15 words or shorter.\newline
The response must not close the dialog.\newline
The response must only include entities of type - choice.\newline
The response must not include any entities of type - name, address, phone, food, area, postcode, price range, type, reference number, stars.\newline
\newline
[dialog history 1]\newline
user: hello i am looking for a place to go , can you help me ?\newline
\newline
[follow-up response 1]\newline
Let's think step-by-step.\newline
As an expert, I must understand the user's requirements from [dialog history 1], identify the relevant information from the [database 1], follow all the [rules 1] and write the response.\newline
I will include entities of type ['choice'] in my response.\newline
I will include these entities - \textcolor{ao(english)}{[('choice', '79')]\newline
assistant: i have found 79 place for you . do you have any specific ideas in mind ?}\newline
\newline
[example 2]\newline
[database 2]\newline
\{\newline
\hspace*{5pt}  \"acorn guest house":\{\newline
\hspace*{10pt}      "address":"154 chesterton road",\newline
\hspace*{10pt}      "phone":"01223353888",\newline
\hspace*{10pt}      "area":"north",\newline
\hspace*{10pt}      "postcode":"cb41da",\newline
\hspace*{10pt}      "price range":"moderate",\newline
\hspace*{10pt}      "type":"guesthouse",\newline
\hspace*{10pt}      "stars":"4 star",\newline
\hspace*{10pt}      "choice":"24"\newline
\hspace*{5pt}    \},....\newline
\}\newline
\newline
[rules 2]\newline
The response must be 10 words or shorter.\newline
The response must not close the dialog.\newline
The response must only include entities of type - choice.\newline
The response must not include any entities of type - name, address, phone, food, area, postcode, price range, type, reference number, stars.\newline
\newline
[dialog history 2]\newline
user: i ' d like to find a guesthouse to stay .\newline
\newline
[follow-up response 2]\newline
Let's think step-by-step.\newline
As an expert, I must understand the user's requirements from [dialog history 2], identify the relevant information from the [database 2], follow all the [rules 2] and write the response.\newline
I will include entities of type ['choice'] in my response.\newline
I will include these entities - \textcolor{ao(english)}{[('choice', '24')]\newline
assistant: no problem . we have 24 to choose from . any specifics ?}\newline
\newline
[example 3]\newline
[database 3]\newline
\{\newline
\hspace*{5pt} "great saint mary ' s church":\{\newline
\hspace*{10pt}     "address":"market square",\newline
\hspace*{10pt}     "phone":"01223350914",\newline
\hspace*{10pt}     "area":"centre",\newline
\hspace*{10pt}     "postcode":"cb23pq",\newline
\hspace*{10pt}     "price range":"cheap",\newline
\hspace*{10pt}     "type":"architecture",\newline
\hspace*{10pt}     "choice":"a lot"\newline
\hspace*{5pt}   \},....\newline
\}\newline
\newline
[rules 3]\newline
The response must be 15 words or shorter.\newline
The response must not close the dialog.\newline
The response must only include entities of type - choice.\newline
The response must not include any entities of type - name, address, phone, food, area, postcode, price range, type, reference number, stars.\newline
\newline
[dialog history 3]\newline
user: i am looking for a place to go !\newline
\newline
[follow-up response 3]\newline
Let's think step-by-step.\newline
As an expert, I must understand the user's requirements from [dialog history 3], identify the relevant information from the [database 3], follow all the [rules 3] and write the response.\newline
I will include entities of type ['choice'] in my response.\newline
I will include these entities -
\end{tcolorbox}

\clearpage

\subsection*{SMD}\label{prompt:smd}
\begin{tcolorbox}[
    colback=white,colframe=black,boxrule=0.25mm,
    width=\textwidth,
]
\fontsize{6pt}{6pt}\selectfont
Henceforth, assume that you are an expert in in-car infotainment. I will give you an incomplete dialog between a user and an in-car infotainment system. As an expert, you must suggest the most appropriate follow-up response to the dialog. Ensure you also include correct information (entities) from the given database. Entities can be of the following types - \newline
1. poi - name of a point of interest, e.g., home, starbucks, pizza chicago, etc.\newline
2. address - address of a poi, e.g, 783 arcadia pl.\newline
3. poi type - the type of a poi, e.g., tea or coffee place, hospital, shopping center, etc.\newline
4. traffic info - traffic status on the way to a poi, e.g., heavy traffic, no traffic, road block nearby, etc. \newline
5. distance - distance of a poi from the user's current location, e.g., 2 miles, 4 miles, etc.\newline
6. event - an event in the user's calendar\newline
7. date - date in a month like the 1st or the 4th or day of a week like monday, wednesday.\newline
8. time - the time on which an event is scheduled\newline
9. party - party attending an event, e.g., tom, boss, brother, executive team, etc.\newline
10. agenda - agenda associated with an event, e.g., discuss dress code, go over budget, etc.\newline
11. room - meeting place of an event, e.g., conference room 100, etc.\newline
12. location - a location for which the user may request the weather information, e.g, boston, los angeles, etc.\newline
13. weather attribute - weather description in a location, e.g., cloudy, warm, hot, overcast etc.\newline
14. temperature - the in a location, e.g., 60f, 100f, etc.\newline
15. weekly time - temporal indicators like today, tomorrow, next week etc.\newline
\newline
As an expert, you are very strict about following rules. Make sure that the follow-up response you write follows all the given rules. Here are the examples -\newline
\newline
[example 1]\newline
[database 1]\newline
\{\newline
\hspace*{5pt}  "trader joes":\{\newline
\hspace*{10pt}    "address":"408 university ave",\newline
\hspace*{10pt}    "poi type":"grocery store",\newline
\hspace*{10pt}    "traffic info":"no traffic",\newline
\hspace*{10pt}    "distance":"5 miles"\newline
\hspace*{5pt}  \},....\newline
\}\newline
\newline
[rules 1]\newline
The response must be 11 words or shorter.\newline
The response must not close the dialog.\newline
The response must only include entities of type - poi, poi type.\newline
The response must not include any entities of type - address, traffic info, distance, event, date, time, party, agenda, room, location, weather attribute, temperature, weekly time.\newline
\newline
[dialog history 1]\newline
user: give me directions to the nearest grocery store\newline
\newline
[follow-up response 1]\newline
Let's think step-by-step.\newline
As an expert, I must understand the user's requirements from [dialog history 1], identify the relevant information from the [database 1], follow all the [rules 1] and write the response.\newline
I will include entities of type ['poi', 'poi type'] in my response.\newline
I will include these entities - \textcolor{ao(english)}{[('poi', 'trader joes'), ('poi type', 'grocery store')]\newline
system: the nearest grocery store is trader joes , would you like directions ?}\newline
\newline
[example 2]\newline
[database 2]\newline
\{\newline
\hspace*{5pt}  "safeway":\{\newline
\hspace*{10pt}    "address":"452 arcadia pl",\newline
\hspace*{10pt}    "poi type":"grocery store",\newline
\hspace*{10pt}    "traffic info":"heavy traffic",\newline
\hspace*{10pt}    "distance":"4 miles"\newline
\hspace*{5pt}  \},....\newline
\}\newline
\newline
[rules 2]\newline
The response must be 23 words or shorter.\newline
The response must not close the dialog.\newline
The response must only include entities of type - distance, poi, traffic info.\newline
The response must not include any entities of type - address, poi type, event, date, time, party, agenda, room, location, weather attribute, temperature, weekly time.\newline
\newline
[dialog history 2]\newline
user: give me directions to the closest grocery store\newline
\newline
[follow-up response 2]\newline
Let's think step-by-step.\newline
As an expert, I must understand the user's requirements from [dialog history 2], identify the relevant information from the [database 2], follow all the [rules 2] and write the response.\newline
I will include entities of type ['distance', 'poi', 'traffic info'] in my response.\newline
I will include these entities - \textcolor{ao(english)}{[('distance', '4 miles'), ('poi', 'safeway'), ('traffic info', 'heavy traffic')]\newline
system: we ' re 4 miles away from safeway but there is heavy traffic in this moment : do i set the gps to go there ?}\newline
\newline
[example 3]\newline
[database 3]\newline
\{\newline
\hspace*{5pt}  "sigona farmers market":\{\newline
\hspace*{10pt}    "address":"638 amherst st",\newline
\hspace*{10pt}    "poi type":"grocery store",\newline
\hspace*{10pt}    "traffic info":"no traffic",\newline
\hspace*{10pt}    "distance":"4 miles"\newline
\hspace*{5pt}  \},....\newline
\}\newline
\newline
[rules 3]\newline
The response must be 10 words or shorter.\newline
The response must not close the dialog.\newline
The response must only include entities of type - distance, poi, poi type.\newline
The response must not include any entities of type - address, traffic info, event, date, time, party, agenda, room, location, weather attribute, temperature, weekly time.\newline
\newline
[dialog history 3]\newline
user: give me directions to the closest grocery store\newline
\newline
[follow-up response 3]\newline
Let's think step-by-step.\newline
As an expert, I must understand the user's requirements from [dialog history 3], identify the relevant information from the [database 3], follow all the [rules 3] and write the response.\newline
I will include entities of type ['distance', 'poi', 'poi type'] in my response.\newline
I will include these entities -
\end{tcolorbox}

\clearpage
\subsection*{BiTOD}\label{prompt:BiTOD}
\begin{tcolorbox}[
    colback=white,colframe=black,boxrule=0.25mm,
    width=\textwidth,
]
\fontsize{5.5pt}{5pt}\selectfont
Henceforth, assume that you are a customer support expert. I will give you an incomplete dialog between a user and a customer service representative. As an expert, you must suggest the most appropriate follow-up response to the dialog. Ensure you also include correct information (entities) from the given database. Entities can be of the following types -\newline
1. name - name of a place (restaurant, hotel or attraction)\newline
2. address - address of the place\newline
3. phone number - phone number of the place\newline
4. location - a part of the city e.g. canal road, central district\newline
5. rating - user rating of the place out of 10 e.g. 8, 9\newline
6. price level - price range of the place, e.g. cheap, moderate, expensive\newline
7. reference number - reference code for booking, e.g. 542j9wog\newline
8. stars - star rating of the hotel, e.g. 3 stars\newline
9. price per night - hotel charges per night e.g. 512, 600, etc.\newline
10. number of rooms - number of rooms to book for the customer e.g. 1, 2\newline
11. number of nights - number of nights the customer wants to book the hotel e.g. 2, 3\newline
12. user name - name of the user e.g. Jack, Henry\newline
13. start month - starting month of the booking e.g. July, May, etc.\newline
14. start day - starting day of the booking e.g. 12, 30 etc.\newline
15. cuisine - the cuisine of a restaurant, e.g. thai, chinese, etc.\newline
16. dietary restrictions - dietary restrictions that the restaurant facilitates e.g. vegan, gluten free\newline
17. number of people - number of people to reserve a restaurant for e.g. 2, 10\newline
18. month - a month of the year e.g. january, february, etc.\newline
19. day - a day of the month/week e.g. 12, 17, monday, etc.\newline
20. time - a time of the daay e.g. 1200, 1330, 930, etc.\newline
21. type - type of an attraction e.g. zoos and aquariums, shopping, etc.\newline
\newline
As an expert, you are very strict about following rules. Make sure that the follow-up response you write follows all the given rules. Here are the examples -\newline
\newline
[example 1]\newline
[database 1]\newline
{\newline
\hspace*{5pt} "house 1881":{\newline
\hspace*{10pt} "phone number":"852 0071 5353",\newline
\hspace*{10pt} "location":"tsim sha tsui",\newline
\hspace*{10pt} "rating":"8",\newline
\hspace*{10pt} "price level":"expensive",\newline
\hspace*{10pt} "reference number":"swm2n2uu",\newline
\hspace*{10pt} "stars":"5",\newline
\hspace*{10pt} "price per night":"1895",\newline
\hspace*{10pt} "number of rooms":"2"\newline
\hspace*{5pt} },...\newline
}\newline
\newline
[rules 1]\newline
The response must be 20 words or shorter.\newline
The response must not close the dialog.\newline
The response must only include entities of type - name, rating.\newline
The response must not include any entities of type - address, phone number, location, price level, reference number, stars, price per night, number of rooms, number of nights, user name, start month, start day, cuisine, dietary restrictions, number of people, month, day, time, type.\newline
\newline
[dialog history 1]\newline
user: hi , i'm looking for recommendations for hotels .\newline
assistant: sure , what hotel rating are you looking for ?\newline
user: i want at least a rating of 1 .\newline
assistant: ok , what about price level or location ?\newline
user: i want a expensive hotel . i don't care about the location .\newline
assistant: sounds good , what about the number of stars of the hotel ?\newline
user: should be at least 2 stars .\newline
\newline
[follow-up response 1]\newline
Let's think step-by-step.\newline
As an expert, I must understand the user's requirements from [dialog history 1], identify the relevant information from the [database 1], follow all the [rules 1] and write the response.\newline
I will include entities of type ['name', 'rating'] in my response.\newline
I will include these entities - \textcolor{ao(english)}{[['name', 'house 1881'], ['rating', '8']]\newline
assistant: i found \#16 hotels that are available . the house 1881 looks good . it has a rating of 8 .}\newline
\newline
[example 2]\newline
[database 2]\newline
{\newline
\hspace*{10pt} "house 1881":{\newline
\hspace*{10pt} "phone number":"852 0071 5353",\newline
\hspace*{10pt} "location":"tsim sha tsui",\newline
\hspace*{10pt} "rating":"8",\newline
\hspace*{10pt} "price level":"expensive",\newline
\hspace*{10pt} "reference number":"swm2n2uu",\newline
\hspace*{10pt} "stars":"5",\newline
\hspace*{10pt} "price per night":"1895",\newline
\hspace*{10pt} "number of rooms":"2"\newline
\hspace*{5pt} },...\newline
}\newline
\newline
[rules 2]\newline
The response must be 19 words or shorter.\newline
The response must not close the dialog.\newline
The response must only include entities of type - name, rating.\newline
The response must not include any entities of type - address, phone number, location, price level, reference number, stars, price per night, number of rooms, number of nights, user name, start month, start day, cuisine, dietary restrictions, number of people, month, day, time, type.\newline
\newline
[dialog history 2]\newline
user: hey ! i am looking for hotels with at least 2 stars . do you have any recommendations ?\newline
assistant: glad to be of service . to get started , can you tell me what rating level and price range are you looking for ?\newline
user: oh yeah ! i am looking for an expensive hotel with minimum 4 rating .\newline
\newline
[follow-up response 2]\newline
Let's think step-by-step.\newline
As an expert, I must understand the user's requirements from [dialog history 2], identify the relevant information from the [database 2], follow all the [rules 2] and write the response.\newline
I will include entities of type ['name', 'rating'] in my response.\newline
I will include these entities - \textcolor{ao(english)}{[['name', 'house 1881'], ['rating', '8']]\newline
assistant: as per your needs , there are \#16 hotels available . i would recommend house 1881 with rating 8 .}\newline
\newline
[example 3]\newline
[database 3]\newline
{\newline
\hspace*{5pt} "jw marriott hotel hong kong":{\newline
\hspace*{10pt} "phone number":"852 7885 6633",\newline
\hspace*{10pt} "location":"hong kong island",\newline
\hspace*{10pt} "rating":"9",\newline
\hspace*{10pt} "price level":"expensive",\newline
\hspace*{10pt} "reference number":"s5y9h2s3",\newline
\hspace*{10pt} "stars":"5",\newline
\hspace*{10pt} "price per night":"2210",\newline
\hspace*{10pt} "number of rooms":"10"\newline
\hspace*{5pt} },...\newline
}\newline
\newline
[rules 3]\newline
The response must be 13 words or shorter.\newline
The response must not close the dialog.\newline
The response must only include entities of type - name, rating.\newline
The response must not include any entities of type - address, phone number, location, price level, reference number, stars, price per night, number of rooms, number of nights, user name, start month, start day, cuisine, dietary restrictions, number of people, month, day, time, type.\newline
\newline
[dialog history 3]\newline
user: hello . i'm trying to find a hotel for my stay with at least 4 stars . would you be able to help me ?\newline
assistant: hi there . i would be happy to help . would you like an expensive or affordable priced hotel ? do you prefer a high rating hotel ?\newline
user: i would like an expensive hotel with a rating of at least 4 .\newline
\newline
[follow-up response 3]\newline
Let's think step-by-step.\newline
As an expert, I must understand the user's requirements from [dialog history 3], identify the relevant information from the [database 3], follow all the [rules 3] and write the response.\newline
I will include entities of type ['name', 'rating'] in my response.\newline
I will include these entities -\newline
\end{tcolorbox}

\end{document}